\documentclass[sigconf,screen]{acmart}

\copyrightyear{2021}
\acmYear{2021}
\setcopyright{acmlicensed}\acmConference[MM '21]{Proceedings of the 29th ACM International Conference on Multimedia}{October 20--24, 2021}{Virtual Event, China}
\acmBooktitle{Proceedings of the 29th ACM International Conference on Multimedia (MM '21), October 20--24, 2021, Virtual Event, China}
\acmPrice{15.00}
\acmDOI{10.1145/3474085.3475521}
\acmISBN{978-1-4503-8651-7/21/10}

\acmSubmissionID{1892}

\usepackage{svg}
\usepackage{amsmath}
\usepackage{footmisc}

\usepackage{mathrsfs}
\usepackage{adjustbox}
\usepackage{tabularx}
\usepackage{booktabs}
\usepackage{multirow}
\usepackage{flushend}
\usepackage{breqn} 
\usepackage{mathtools}

\newcolumntype{m}[1]{>{\centering\arraybackslash}p{#1}}
\usepackage{subcaption}
\usepackage{balance}

\definecolor{OliveGreen}{rgb}{0,0.6,0}
\definecolor{SoftRed}{rgb}{1,0.2,0.2}
\usepackage{xcolor}
\newcommand{\capitalhyphen}{\raisebox{0.24ex}{\resizebox{0.4em}{\height}{-}}\kern-0.07em}
\usepackage{color}
\usepackage{caption}
\usepackage{epsfig}
\usepackage{enumitem}
\usepackage{mathtools}
\usepackage{epsfig}
\usepackage{graphicx}
\usepackage{breqn} 
\usepackage{placeins} 

\newcommand{\link}[1]{{\color{blue}\href{#1}{#1}}}

\begin{document}
\fancyhead{}
\title[MeronymNet]{MeronymNet: A Hierarchical Approach for Unified and Controllable Multi-Category Object Generation}

\author{Rishabh Baghel}
\email{baghelrishabha@gmail.com}
\affiliation{%
  \institution{CVIT, IIIT Hyderabad}
  \city{Hyderabad 500032}
  \country{INDIA}
}
\author{Abhishek Trivedi}
\email{abhishek.trivedi@research.iiit.ac.in}
\affiliation{%
  \institution{CVIT, IIIT Hyderabad}
  \city{Hyderabad 500032}
  \country{INDIA}
}
\author{Tejas Ravichandran}
\email{venkatakrishnan164@gmail.com}
\affiliation{%
  \institution{CVIT, IIIT Hyderabad}
  \city{Hyderabad 500032}
  \country{INDIA}
}
\author{Ravi Kiran Sarvadevabhatla}
\email{ravi.kiran@iiit.ac.in}
\affiliation{%
 \institution{CVIT, IIIT Hyderabad}
 \city{Hyderabad 500032}
 \country{INDIA}
 }

\renewcommand{\shortauthors}{Baghel et al.}

\begin{abstract}
  We introduce MeronymNet, a novel hierarchical approach for controllable, part-based generation of multi-category objects using a single unified model. We adopt a guided coarse-to-fine strategy involving semantically conditioned generation of bounding box layouts, pixel-level part layouts and ultimately, the object depictions themselves. We use Graph Convolutional Networks, Deep Recurrent Networks along with custom-designed Conditional Variational Autoencoders to enable flexible, diverse and category-aware generation of 2-D objects in a controlled manner. The performance scores for generated objects reflect MeronymNet's superior performance compared to multiple strong baselines and ablative variants. We also showcase MeronymNet's suitability for controllable object generation and interactive object editing at various levels of structural and semantic granularity.
\end{abstract}



\begin{teaserfigure}
   \centering
     \includegraphics[width=\textwidth]{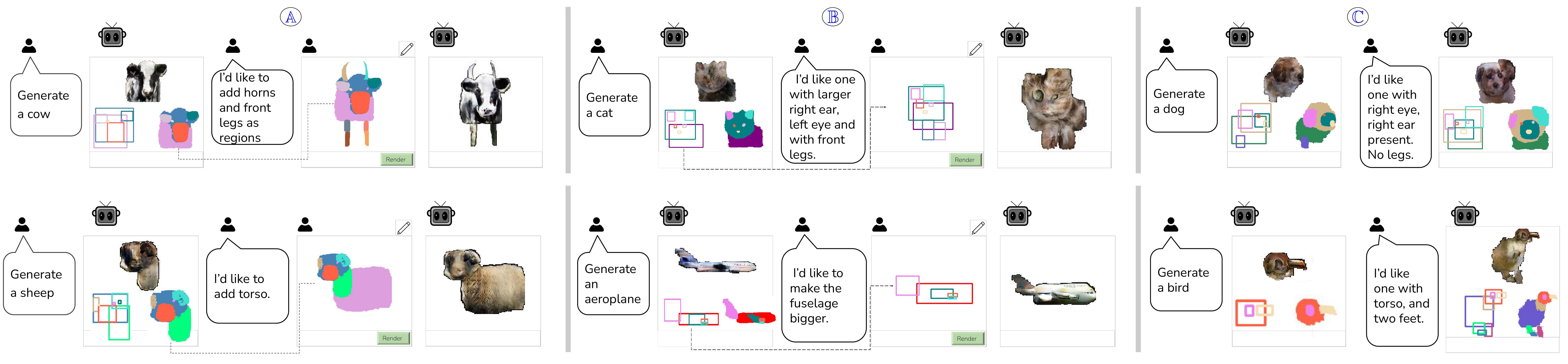}
     \captionof{figure}{Wireframe mockups of our interactive app \textsc{MeroBot}'s user interface depicting object representations generated by MeronymNet.  The first frame of panels in $\mathbb{A,B,C}$ shows the generated part bounding box (bottom-left), pixel level label map (bottom-right) and color object depiction (top). As the panels illustrate, our approach enables part-based object generations to be interactively modified at part region level ($\mathbb{A}$), bounding boxes ($\mathbb{B}$) and at part list level ($\mathbb{C}$). Panel $\mathbb{B}$ depicts a cartoon-like scenario where parts are exaggerated in size by the user. The pencil icon indicates the canvas contents can be edited by user. The icons at top-left corner of each frame refer to active entity (user or \textsc{MeroBot}). Refer to Sec.~\ref{sec:interactive}.}
     \label{fig:merobot}
\end{teaserfigure}

\maketitle

\section{Introduction}

Many recent successes for generative approaches have been associated with generation of realistic scenery using top-down guidance from text~\cite{hong2018inferring} and semantic scene attributes~\cite{ritchie2019fast,zhaobo2019layout2im,Sun_2019_ICCV,park2019SPADE,Azadi2019SemanticBS}. Alongside, generative models for individual objects have also found a degree of success~\cite{bessinger2019generative,he2019attgan,cheng2020segvae,yin2019semantics,zhang2018stackgan++,brock2018large}. Despite these successes, the degree of controllability in these approaches is somewhat limited compared to the scene generation counterparts. 

The bulk of object generation approaches specialize for a single category of objects with weakly aligned part configurations (e.g. faces~\cite{bessinger2019generative,he2019attgan,cheng2020segvae}) and for objects with associated text description (e.g. birds~\cite{yin2019semantics,zhang2018stackgan++}). Although models such as as BigGAN~\cite{brock2018large} go beyond a single category, conditioning the generative process is possible only at a category level. In addition, all of the existing approaches induce contextual bias by \textit{necessarily} generating background along with the object. As a result, the generated object cannot be utilized as a sprite (mask) for downstream tasks~\cite{charity2020baba} (e.g. compositing the object within a larger scene). 

To address the shortcomings mentioned above, we introduce MeronymNet\footnote{Meronym is a linguistic term for expressing part-to-whole relationships.}, a novel unified part and category-aware generative model for object sprites (Fig.~\ref{fig:overview}). Conditioned on a specified object category and an associated part list, a first-level generative model stochastically generates a part bounding box layout (Sec. \ref{sec:BoxGCN-VAE}). The category  and generated part bounding boxes are used to guide a second-level model which generates semantic part maps (Sec.~\ref{sec:LabelMap-VAE}). Finally, the semantic part label maps are conditionally transformed into object depictions via the final level model (Sec.~\ref{sec:label2obj}). Our unified approach enables diverse, part-controllable object generations across categories using a \textit{single} model, without the necessity of multiple category-wise models. Additionally, MeronymNet's design enables interactive generation and editing of objects at multiple levels of semantic granularity (Sec.~\ref{sec:interactive}). Please visit our project page \link{http://meronymnet.github.io/} for source code, generated samples and additional material.

\section{Related Work}

\noindent \textbf{2-D object generative models:} Within the 2-D realm, approaches are designed for a specific object type (e.g. faces~\cite{bessinger2019generative, cheng2020segvae}, birds~\cite{yin2019semantics,zhang2018stackgan++}, flowers~\cite{park2018mc}) and usually involve text or pixel-level conditioning~\cite{isola2017image}. In some cases, part attributes are used to generate these specific object types~\cite{he2019attgan,yan2016attribute2image}. BigGAN~\cite{brock2018large} is a notable example for multi-category object generation. Apart from these works, objects are usually generated as intermediate component structures by scene generation approaches~\cite{hong2018inferring,park2019SPADE,Jyothi_2019_ICCV}. Unified, part-controllable multi-category object generative models do not exist, to the best of our knowledge. 

\noindent \textbf{3-D object generative models:} A variety of part-based generative approaches exist for 3-D objects~\cite{SAGnet19,Mo_2019_CVPR,mo2019structurenet,li2017grass,nash2017shape,Wu_2020_CVPR}. Unlike our unified model, these approaches generally train a separate model for each object. Also, the number of object categories, maximum number of parts and variation in intra-category spatial articulation in these approaches is generally smaller compared to our setting.

\noindent \textbf{Scene Generation:} A common paradigm is to use top-down guidance from textual scene descriptions to guide a two stage approach~\cite{hong2018inferring,Jyothi_2019_ICCV}. The first stage produces object bounding box layouts while the second stage component~\cite{zhaobo2019layout2im,Sun_2019_ICCV} translates these layouts to scenes. In some cases, the first stage produces semantic maps~\cite{cheng2020segvae,Azadi2019SemanticBS} which are translated to scenes using pixel translation models~\cite{isola2017image,park2019SPADE}. We too employ the approach of intermediate structure generations but with three levels in the generative hierarchy and with object attribute conditioning.    

 \noindent \textbf{Graph Convolutional Networks (GCNs):} GCNs have emerged as a popular framework for working with graph-structured data and predominantly for discriminative tasks~\cite{zhangdual,verma2018feastnet,yang2019auto,zhao2019semantic}. In a generative setting, GCNs have been utilized for scene graph generation~\cite{yang2018graph,gu2019scene}. In our work, we use GCNs which operate on part-graph object representations. These representations, in turn, are used to optimize a Variational Auto Encoder (VAE) generative model. GCN--VAE have been used for modelling citation, gene interaction networks~\cite{kipf2016variational,yang2018meta} and molecular design~\cite{liu2018constrained}. To the best of our knowledge, we are the first to introduce a conditional variant of GCN--VAE for multi-category object generation. 
 
\section{MeronymNet}
\label{sec:meronymnet}

We begin with a brief overview of Variational Auto Encoder (VAE)~\cite{kingma2013auto} and its extension Conditional Variational Auto Encoder (CVAE)~\cite{sohn2015learning}. Subsequently, we provide a summary of our overall generative pipeline (Sec. \ref{sec:approachsummary}). The sections following the summary describe the components of MeronymNet in detail (Sec.~\ref{sec:BoxGCN-VAE}-~\ref{sec:label2obj}). 

\begin{figure}[!t]
  \centering
  \includegraphics[width=\linewidth]{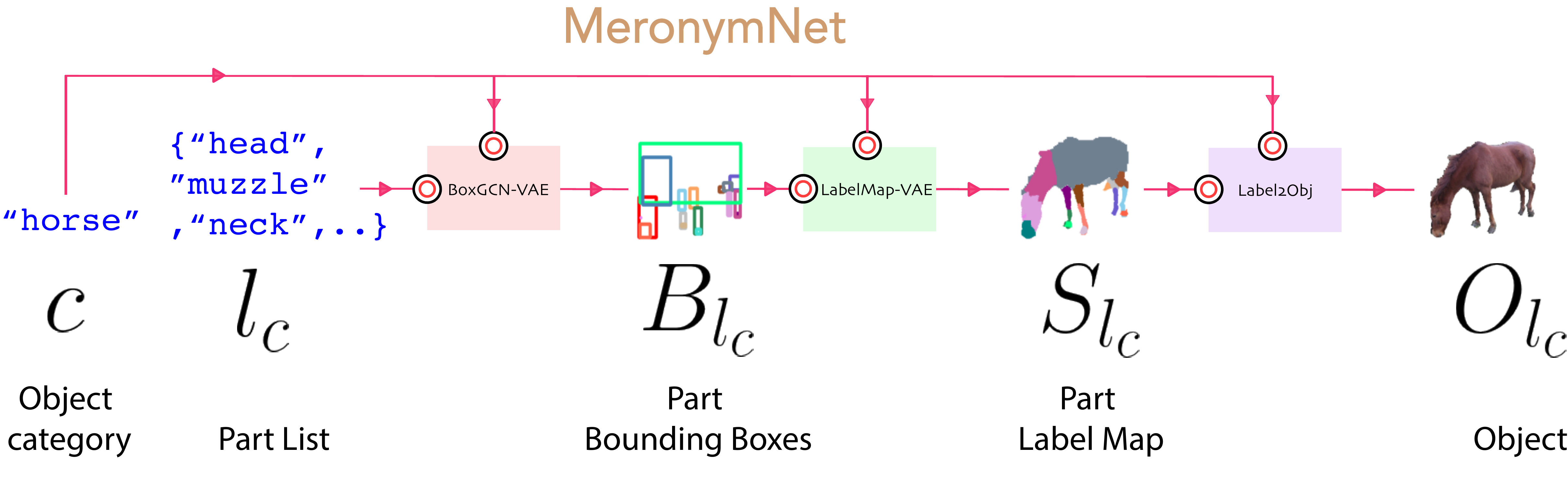}
  \caption{Given an object category $c$ and an associated list of plausible object parts $l_c$, a part-labelled bounding box layout ${B}_{l_c}$ is stochastically generated (Sec.~\ref{sec:BoxGCN-VAE}). The box layout and category is used to guide the generation of a pixel-level label map $S_{l_c}$ (Sec.~\ref{sec:LabelMap-VAE}). This layout map is translated into the final object depiction $O_{l_c}$ (Sec.~\ref{sec:label2obj}). Black-red circles indicate conditioning by object attributes during generation.}
  \label{fig:overview}
\end{figure}

\subsection{VAE and CVAE}
\label{sec:vae}

\noindent \textbf{VAE:} Let $p(\mathsf{d})$ represent the probability distribution of data. VAEs transform the problem of generating samples from $p(\mathsf{d})$ into that of generating samples from the likelihood distribution $p_{\theta}(\mathsf{d}|z)$ where $z$ is a low-dimensional latent surrogate of $\mathsf{d}$, typically modelled as a standard normal distribution, i.e. $p(z) = \mathcal{N}(0,I)$. Obtaining a representative latent embedding for a given data sample $\mathsf{d}$ is possible if we have the exact posterior distribution  $p(z|\mathsf{d})$. Since the latter is intractable, a variational approximation $q_{\phi}(z|\mathsf{d})$, which is easier to sample from, is used. The distribution is modelled using an encoder neural network parameterized by $\phi$. Similarly, the likelihood distribution is modelled by a decoder neural network parameterized by $\theta$. To jointly optimize for $\theta$ and $\phi$, the so-called evidence lower bound (ELBO) of the data's log distribution ($\text{log } p(\mathsf{d})$) is sought to be maximized. The ELBO is given by $\mathcal{L}(\mathsf{d};\phi,\theta) = \mathbb{E}_{q_{\phi}(z|\mathsf{d})} [\text{log } p_{\theta}(\mathsf{d}|z)] - \lambda KL(q_{\phi}(z|\mathsf{d})\lvert \lvert p(z))$ where KL stands for KL-divergence and $\lambda >0$ is a tradeoff hyperparameter.  

\begin{figure*}[!t]
  \centering
  \includegraphics[width=\textwidth]{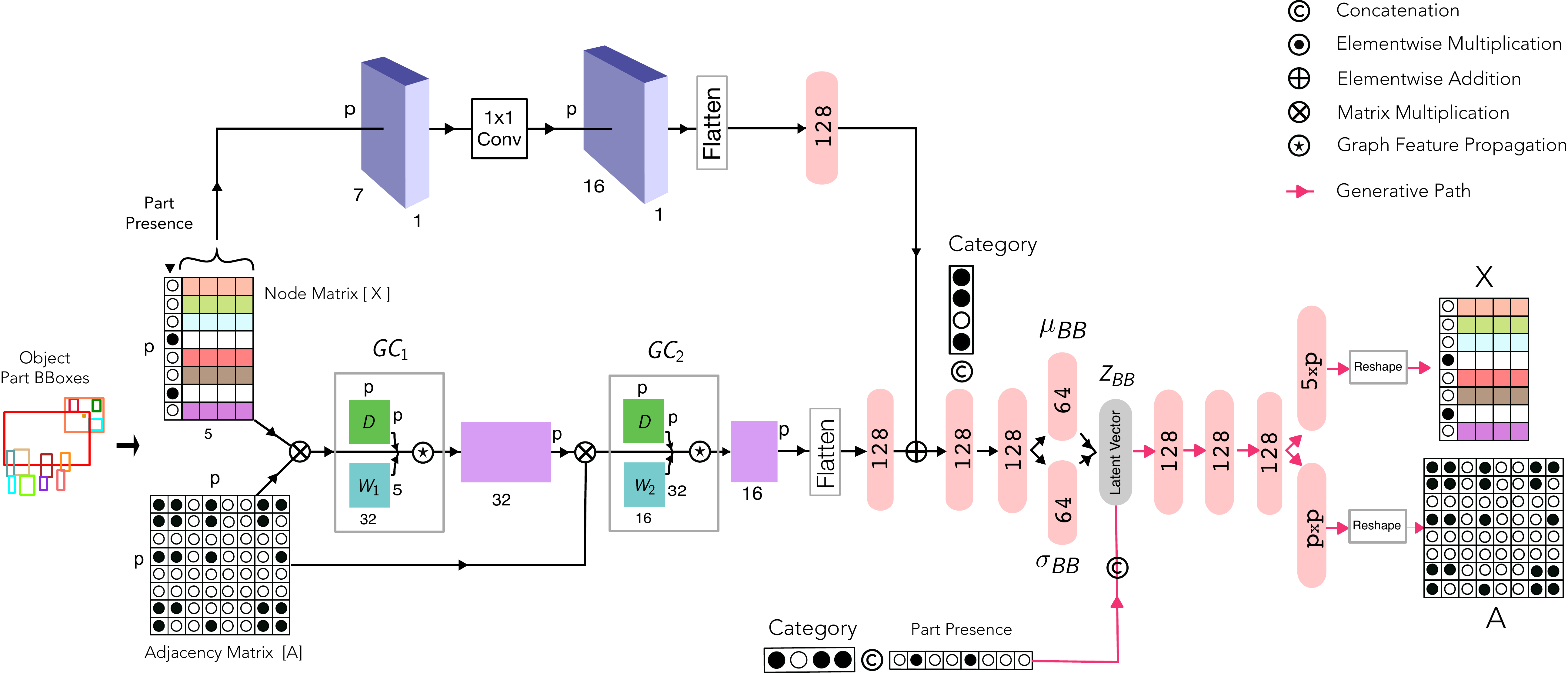}
  \caption{Architecture for BoxGCN--VAE which generates part-labelled bounding box object layouts (Sec.~\ref{sec:BoxGCN-VAE}). The numbers within the rounded light pink rectangles indicate dimensionality of fully connected layers. Generative path is shown in dark pink.}
  \label{fig:merobox}
\end{figure*}

\noindent \textbf{CVAE:} Conditional VAEs are an extension of VAEs which incorporate auxiliary information $\mathsf{a}$ as part of the encoding and generation process. As a result, the decoder now models $p_{\theta}(\mathsf{d}|z,\mathsf{a})$ while the encoder models $q_{\phi}(z|\mathsf{d},\mathsf{a})$. Accordingly, ELBO is modified as $\mathcal{L}(\mathsf{d,a};\phi,\theta) = \mathbb{E}_{q_{\phi}(z|\mathsf{d,a})} [\text{log } p_{\theta}(\mathsf{d}|z,\mathsf{a})]- KL(q_{\phi}(z|\mathsf{d,a})\lvert \lvert p(z|\mathsf{a}))    $

In our problem setting, the conditioning is performed as a gating operation where auxiliary information $\mathsf{a}$ either modulates a feature representation of input (encoder phase) or modulates the latent variable $z$ (decoder phase). We modify CVAE and design an even more general variant where the conditioning used for encoder and decoder can be different. For additional details on VAE and CVAE, refer to the excellent tutorial by Doersch~\cite{doersch2016tutorial}.

\subsection{Summary of our approach}
\label{sec:approachsummary}

Suppose we wish to generate an object from category $c$ $(1 \leqslant c \leqslant M)$, with an associated maximal part list $L_c$. The category ($c$) and a list of parts $l_c \subseteq L_c$ is used to condition BoxGCN--VAE, the first-level generative model (Sec. \ref{sec:BoxGCN-VAE}, pink box in Fig. \ref{fig:overview}) which stochastically generates part-labelled bounding boxes. The generated part-labeled box layout ${B}_{l_c}$ and $c$ are used to condition LabelMap--VAE  (Sec. \ref{sec:LabelMap-VAE}) which generates a category-specific per-pixel part-label map $S_{l_c}$. Finally, the label map is conditionally transformed into an RGB object depiction $O_{l_c}$ via the final-level Label2obj model (Sec.~\ref{sec:label2obj}). Next, we provide architectural details for modules that constitute MeronymNet.

\subsection{BoxGCN--VAE}
\label{sec:BoxGCN-VAE}

\noindent \textbf{Representing the part bounding box layout:} We design BoxGCN--VAE (Fig. \ref{fig:merobox}) as a Conditional VAE which learns to stochastically generate part-labelled bounding boxes. To obtain realistic layouts, it is important to capture the semantic and structural relationships between object parts in a comprehensive manner. To meet this requirement, we model the object layout as an undirected graph. We represent this graph in terms of Feature Matrix $X$ and Adjacency Matrix $A$~\cite{kipf2016variational,zhang2018graph}. Let $\mathsf{p}$ be the maximum possible number of parts across all object categories, i.e. $\mathsf{p} = \max_{c} length(L_c), 1 \leqslant c \leqslant M$. $X$ is a $\mathsf{p} \times 5$ matrix where each row $r$ corresponds to a part. A binary value $p_r \in \{0,1\}$ is used to record the presence or absence of the part in the first column. The next $4$ columns represent bounding box coordinates. For categories with part count less than $\mathsf{p}$ and for absent parts, the rows are filled with $0$s. The $\mathsf{p} \times \mathsf{p}$ binary matrix $A$ encodes the connectivity relationships between the object parts in terms of standard part-of (meronymic) relationships (e.g. \textit{nose} is a part of \textit{face})~\cite{beckwith2021wordnet}. Thus, we obtain the object part bounding box representation $\mathbb{G} = (X,A)$.

\noindent \textbf{Graph Convolutional Network (GCN):} A GCN processes a graph $\mathbb{G}$ as input and computes hierarchical feature representations for each graph node. The feature representation at the ($i+1$)-th layer of the GCN is defined as  $H_{i+1} = f(H_i, A)$ where $H_i$ is a $\mathsf{p} \times F_i$  matrix whose $j$-th row contains the feature representation for node indexed by $j$ ($1 \leqslant j \leqslant \mathsf{p}$). $f$ represents the so-called propagation rule which determines the manner in which node features of the previous layer are aggregated to obtain the current layer's feature representation. We use the following propagation rule~\cite{kipf2017semi}: $f(H_i,A) = \sigma({D}^{\frac{-1}{2}} \widetilde{A} {D}^{\frac{-1}{2}} H_i W_i)$ where $\widetilde{A} = A + I$ represents the adjacency matrix modified to include self-loops, ${D}$ is a diagonal node-degree matrix (i.e. ${D}_{jj} = \sum_k  \widetilde{A}_{jk}$) and $W_i$ are the trainable weights for $i$-th layer. $\sigma$ represents a non-linear activation function. Note that $H_0 = X$ (input feature matrix).

\begin{figure*}[!t]
  \centering
  \includegraphics[width=\textwidth]{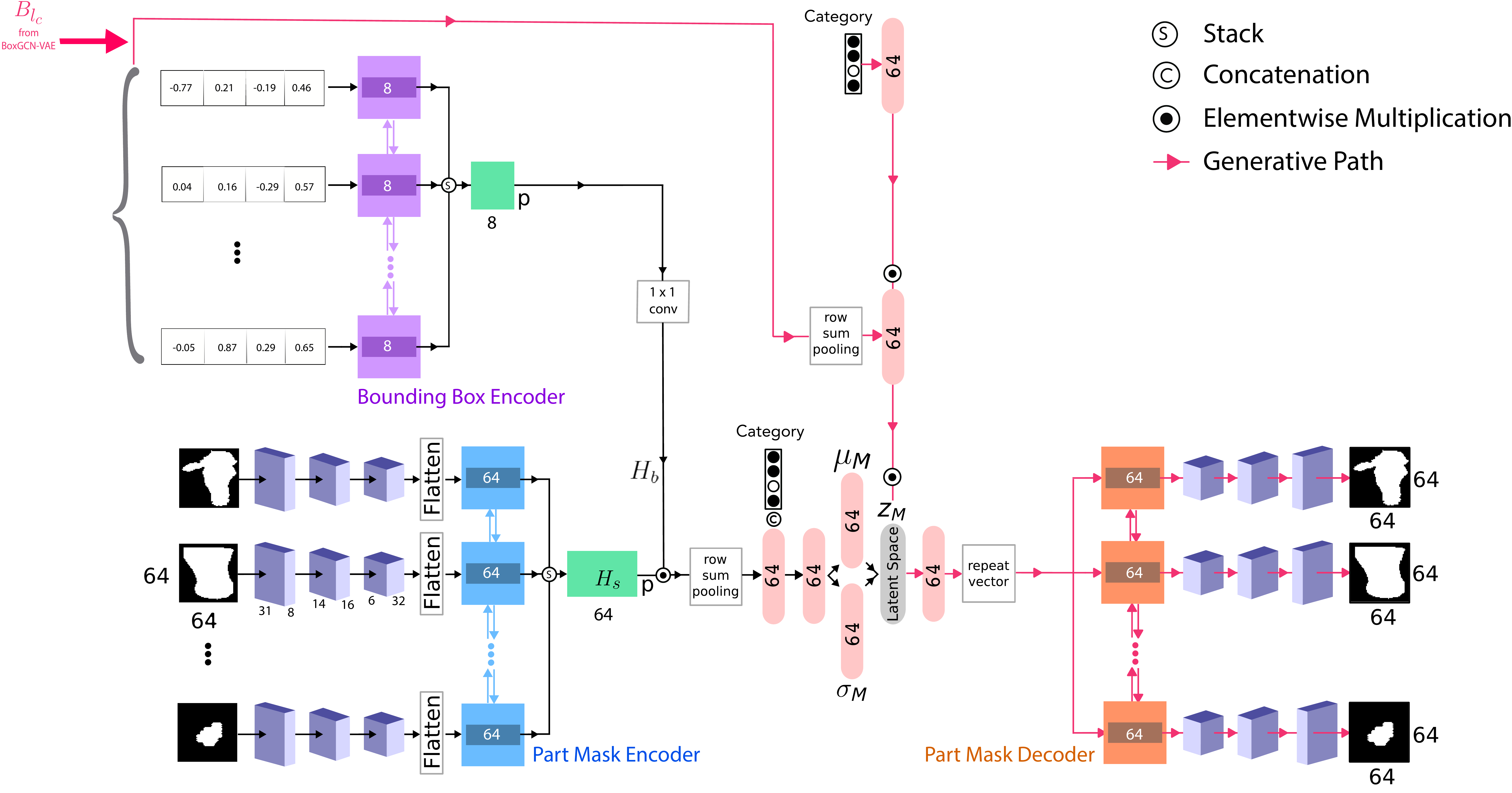}
  \caption{The architecture for LabelMap--VAE which generates per-pixel map for each part conditioned on object class and the part-labeled bounding box layout (Sec.~\ref{sec:LabelMap-VAE}). After generation, the bounding boxes are used to compose the object in terms of appropriately warped part masks. The pink arrows indicate the data flow for the generative model.}
  \label{fig:LabelMap-VAE}
\end{figure*}

\noindent \textbf{Encoding the graph feature representation:} The feature representation of graph $\mathbb{G}$ obtained from GCN is then mapped to the parameters of a diagonal Gaussian distribution ($\mu_{BB}(\mathbb{G}),\sigma_{BB}(\mathbb{G})$), i.e. the approximate posterior. This mapping is guided via category-level conditioning (see Fig.~\ref{fig:merobox}). In addition, the mapping is also conditioned using skip connection features. These skip features are obtained via $1 \times 1$ convolution along the spatial dimension of bounding box sub-matrix of input $\mathbb{G}$ (see top part of Fig. \ref{fig:merobox}). In addition to  part-focused guidance for layout encoding, the skip-connection also helps avoid imploding gradients. 

\noindent \textbf{Reconstruction:} The sampled latent variable $z$, conditioned using category and part presence variables ($c,l_c$), is mapped by the decoder to the components of $\mathbb{G} = (X,A)$. Denoting $X = \left[ X_1 \lvert X_{bb}\right]$, the binary part-presence vector $X_1$ is modelled as a factored multivariate Bernoulli distribution

\begin{align}
p_{{\tiny \theta_{b}}}(X_1|z,c,l_c) =  \prod_{k=1}^{\mathsf{p}} {D_{l_c^k}}^{l_c^k} {(1 - D_{l_c^k})}^{1-l_c^k}    
\end{align}

where $D_{l_c}$ is the corresponding output of the decoder. To encourage accurate localization of part bounding boxes $X_{bb}$, we use two per-box instance-level losses: mean squared error $L_i^{MSE} = \sum_{j=1}^4 (X_{bb}^i[j] - \hat{X}_{bb}^i[j])^2$ and Intersection-over-Union between the predicted ($\hat{X}_{bb}$) and ground-truth ($X_{bb}$) bounding boxes $L_i^{IoU} = -ln (IoU(X_{bb}^i,\hat{X}_{bb}^i))$~\cite{UnitBox2016}. To impose additional structural constraints, we also use a pairwise MSE loss defined over distance between centers of bounding box pairs. Denoting the ground-truth Euclidean distance between centers of $m$-th and $n$-th bounding boxes as $d_{m,n}$, the pairwise loss is defined as $L_{m,n}^{MSE-c} = (d_{m,n} - \hat{d}_{m,n})^2$ where $\hat{d}_{m,n}$ refers to predicted between-center distance.

For the adjacency matrix ($A$), we use binary cross-entropy $L_{m,n}^{BCE}$ as the per-element loss. The overall reconstruction loss for the BoxGCN--VAE decoder for a given object can be written as:

\begin{equation}
\resizebox{.92\hsize}{!}{$\mathcal{L}_{\mathsf{Box}} = \underbrace{\frac{- ln (p_{{\tiny \theta_{b}}}(X_1|z,c,l_c))}{\mathsf{p}}}_{X_1} + \underbrace{\frac{\sum_{i=1}^{\mathsf{p}} ( L_i^{MSE} + L_i^{IoU} )}{\mathsf{p}} +   \frac{\sum_{m=1}^{\mathsf{p}} \sum_{\substack{n=1\\n \neq m}}^{\mathsf{p}} L_{m,n}^{MSE-c}}{\mathsf{p} (\mathsf{p}-1)}}_{X_{bb}} + \underbrace{\frac{\sum_{m=1}^{\mathsf{p}} \sum_{n=1}^{\mathsf{p}} L_{m,n}^{BCE}}{\mathsf{p}^2}}_{A}$}
\label{eqn:BoxGCN-VAE-recons-loss}
\end{equation}

It is important to note that unlike scene graphs~\cite{yang2018graph,gu2019scene}, spatial relationships between nodes (parts) in our object graphs are not explicitly specified. This makes our part graph generation problem setting considerably harder compared to scene graph generation.

Also note that the decoder architecture is considerably simpler compared to encoder. As our experimental results shall demonstrate (Sec. \ref{sec:experiments}), the conditioning induced by category and part-presence, combined with the connectivity encoded in the latent representation $z$, turn out to be adequate for generating the object bounding box layouts despite the absence of graph unpooling layers in the decoder.

\subsection{LabelMap--VAE}
\label{sec:LabelMap-VAE}

Unlike bounding boxes which represent a coarse specification of the object, generating the object label map requires spatial detail for each part to be determined at pixel level. To meet this requirement, we design LabelMap-VAE as a Conditional VAE which learns to stochastically generate per-part spatial masks (Fig.~\ref{fig:LabelMap-VAE}). To guide mask generation in a category-aware and layout-aware manner, we use feature embeddings of object category $c$ and bounding box layout ${B}_{l_c}$ generated by BoxGCN--VAE (Sec.~\ref{sec:BoxGCN-VAE}). 

\noindent \textbf{Conditional encoding of part masks:} During encoding, the binary mask for each part is resized to fixed dimensions. The per-part CNN-encoded feature representations of individual part masks are aggregated using a bi-directional Gated Recurrent Unit (GRU) (color-coded blue in Fig. \ref{fig:LabelMap-VAE}). The per-part hidden-state representations from the unrolled GRU are stacked to form a $\mathsf{p} \times h_s$ representation $H_s$.

\begin{figure}[!t]
  \centering
  \includegraphics[width=\linewidth]{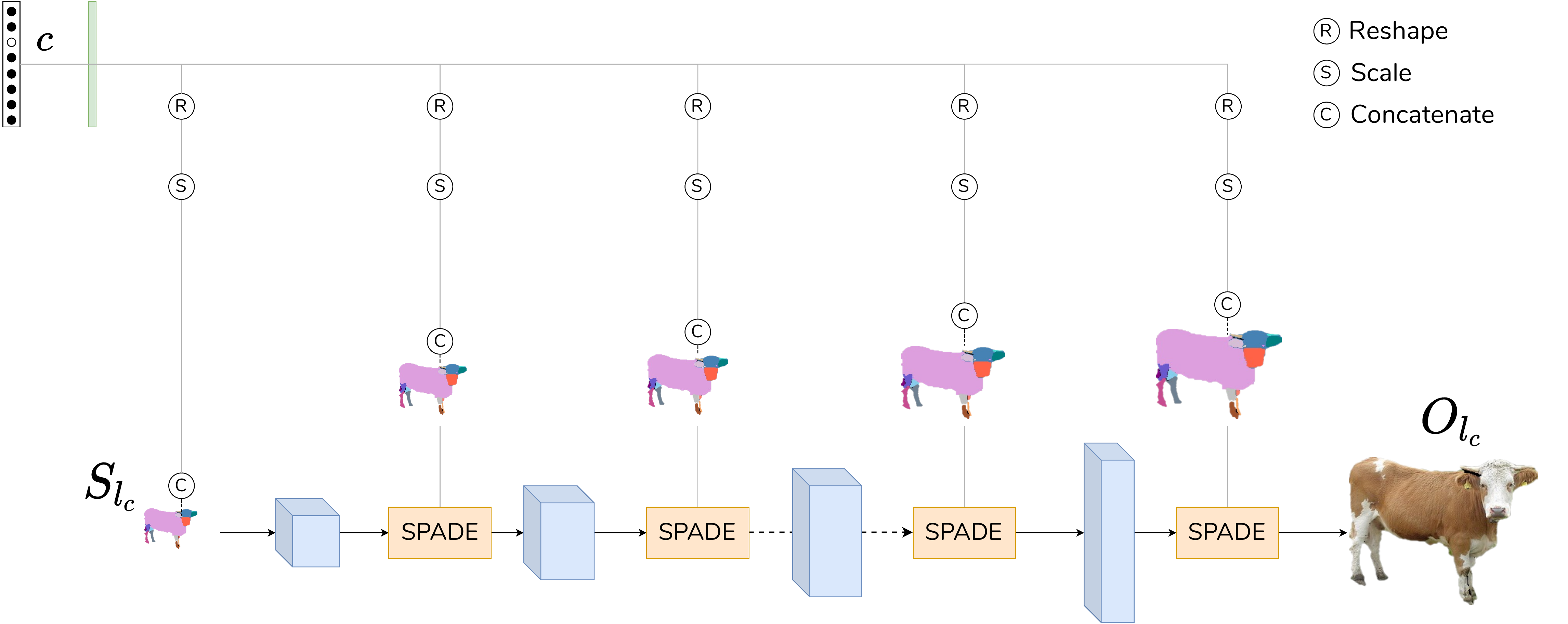}
  \caption{Architecture for Label2obj (Sec. \ref{sec:label2obj}) which translates part label map $S_{l_c}$ to corresponding 2-D object depiction $O_{l_c}$ conditioned on category $c$'s embedding.}
  \label{fig:label2obj}
\end{figure}

In parallel, feature representations for each part bounding box ${B}_{l_c}$ (generated by BoxGCN--VAE) are obtained using another bi-directional GRU (color-coded purple in Fig.~\ref{fig:LabelMap-VAE}). An aggregation and stacking scheme similar to the one used for part masks is used to obtain a $\mathsf{p} \times h_b$ feature matrix. Using $1 \times 1$ convolution, the latter is transformed to a $\mathsf{p} \times h_s$ representation $H_b$ (see Fig.~\ref{fig:LabelMap-VAE}). To induce bounding-box based conditioning, we use $H_b$ to multiplicatively gate the intermediate label-map feature representation $H_s$. The result is pooled across rows and further gated using category information. In turn, the obtained feature representation is ultimately mapped to the parameter representations of a diagonal Gaussian distribution ($\mu_{M},\sigma_{M}$).

\noindent \textbf{Generating the part label map:} The decoder maps the sampled latent variable $z$ to a conditional data distribution $p_{\theta_m}(M|z,\alpha_{bb},\alpha_{c})$ over the sequence of label maps $M = \{m_1,m_2,\ldots,m_p$\} with $\theta_m$ representing parameters of the decoder network. $\alpha_c$ and $\alpha_{bb}$ respectively represent the conditioning induced by feature representations of object category $c$ and the stochastically generated bounding box representation ${B}_{l_c}$. The gated latent vector is mapped to $z_g$ which is replicated $\mathsf{p}$ times and fed to the decoder bi-directional GRU (color-coded orange in Fig. \ref{fig:LabelMap-VAE}). The hidden state of each unrolled GRU unit is subsequently decoded into individual part masks. We model the distribution of part masks as a factored product of conditionals: $p_{\theta_m}(M|z,\alpha_{bb},\alpha_{c}) = \prod_{k=1}^{\mathsf{p}} p(m_k | m_{1:(k-1)}, z, \alpha_{bb}, \alpha_{c})$. In turn, we model each part mask as $p(m_k | m_{1:(k-1)}, z, \alpha_{bb}, \alpha_{c}) = \mathsf{ Softmax}(Z_{\theta}^k)$ where $Z_{\theta}^k$ represents the binary logits from $k$-th part's GRU-CNN decoder.  

To obtain a one-hot representation of the object label map, each part mask is positioned within a $H \times W \times \mathsf{p}$ tensor where $H \times W$ represents the 2-D spatial dimensions of the object canvas. The part's index $k, 1 \leqslant k \leqslant \mathsf{p}$, is used to determine one-hot label encoding while the spatial geometry is obtained by scaling the mask according to the part's bounding box. 

\begin{figure}[!t]
  \centering
  \includegraphics[width=0.9\linewidth]{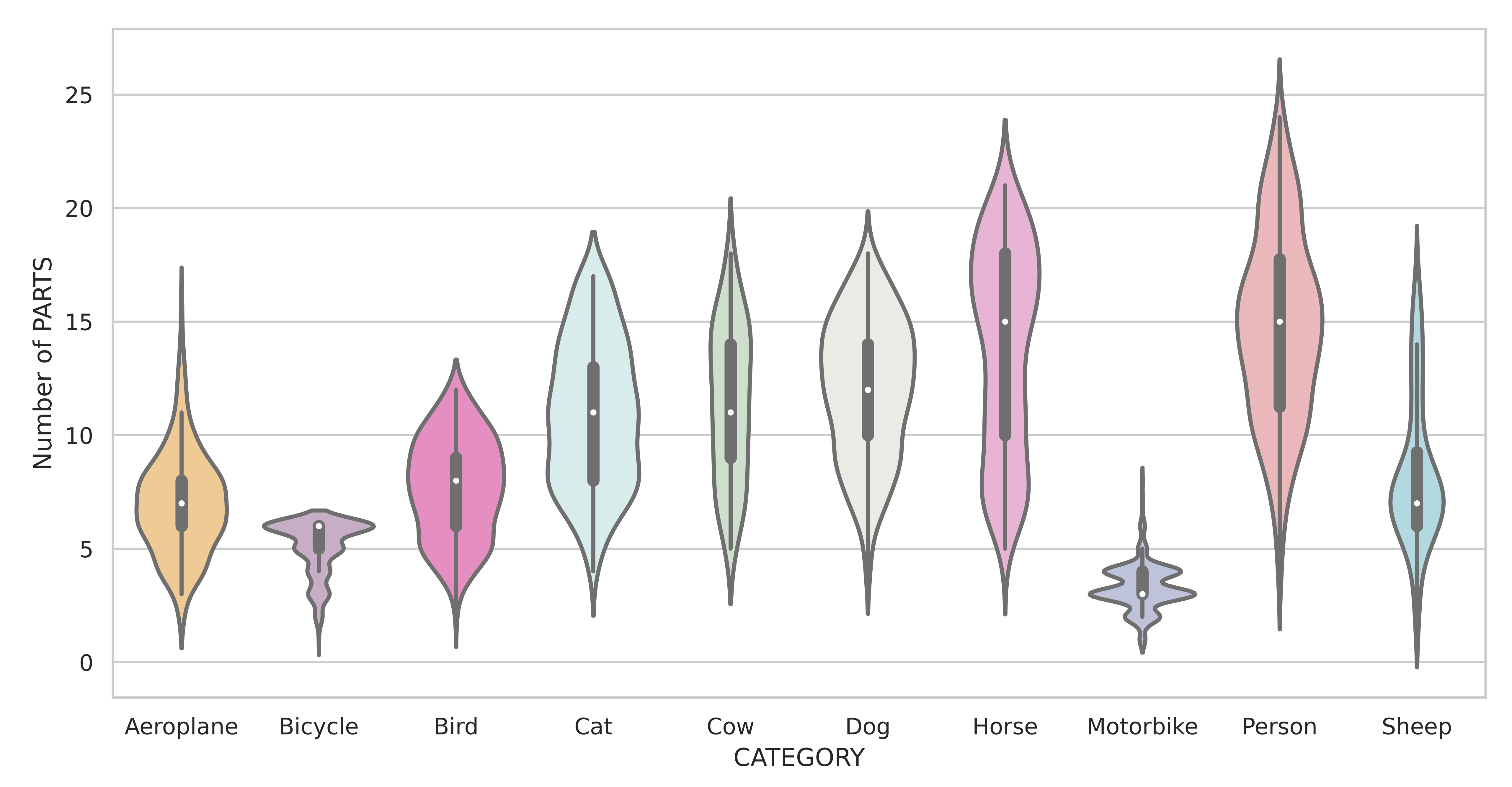}
  \caption{Density distribution of part counts for each object category in the dataset (Sec.~\ref{sec:experiments}) as violin plots.}
  \label{fig:violin}
\end{figure}

\subsection{Label2obj}
\label{sec:label2obj}

For translating part label maps generated by LabelMap--VAE (Sec.~\ref{sec:LabelMap-VAE}) to the final object depiction in a category-aware manner, we design Label2obj as a conditioned variant of the approach by Park et. al.~\cite{park2019SPADE} (see Fig.~\ref{fig:label2obj}). The one-hot representation of object category is transformed via an embedding layer. The embedding output is reshaped, scaled and concatenated depth-wise with appropriately warped label maps. The resultant features comprise one of the inputs to the SPADE blocks within the pipeline. The modified loss functions for Label2obj are:

\bigskip

\begin{dgroup*}
\begin{dmath*}
{L}_{{D}} = -\mathbb{E}_{{l}_{c},c}[\min(0,-1 - {D}_{cS}({l}_{c},G_{cS}({l}_{c},c),c)))] 
\\
- \mathbb{E}_{{l}_{c},{o}_{c},c}[\min(0, -1 + {D}_{cS}({l}_{c}, {o}_{c}, c))] 
\end{dmath*}
\vspace{-5pt}
\begin{dmath*}
    {L}_{{G}} = - \mathbb{E}_{{l}_{c},c} [{D}_{cS}( {l}_{c}, G_{cS}({l}_{c},c),c )] 
       + {\lambda}_{G} {L}_{G}^{feat} + {\lambda}_{D} {L}_{D}^{feat}
\end{dmath*}
\end{dgroup*}

\noindent where ${L}_{{G}}$, ${L}_{{D}}$ are the generator and discriminator losses, ${L}_{G}^{feat}$ and ${L}_{D}^{feat}$, the generator and discriminator feature matching losses~\cite{park2019SPADE}. The subscripts for expectation refer to RGB images ${o}_{c}$, corresponding semantic label map ${l}_{c}$ and object category embeddings $c$. Our modification incorporates a category-aware discriminator~\cite{odena2017conditional} ($D_{cS}$) which complements our category-conditioned generator ($G_{cS}$).

\subsection{Implementation Details}
\label{sec:impldetails}

The VAE-based components (BoxGCN--VAE, LabelMap--VAE) are trained using the standard approach of maximizing the ELBO (Sec. \ref{sec:vae}) and with Adam optimizer~\cite{kingma2014adam}. For the VAE hyperparameter $\lambda$ which trades off reconstruction loss and KL regularization term, we employ a cyclic annealing schedule~\cite{fu2019cyclical}. This is done to mitigate the possibility of the KL term vanishing and to make use of informative latent representations from previous cycles as warm restarts. In addition, we impose a constraint over the difference of training and validation losses. Whenever the loss difference increases above a threshold ($0.1$ in our case), $\lambda$ (KL regularization term) is frozen and prevented from increasing according to the default annealing schedule. BoxGCN--VAE is trained with a learning rate of $10^{-4}$ for $300$ epochs using a mini-batch of size $32$. LabelMap--VAE is trained with a learning rate of $10^{-3}$ for $110$ epochs using a mini-batch size of $8$. The bounding boxes generated by BoxGCN--VAE are scaled to $500 \times 500$ while the label masks generated by LabelMap--VAE are positioned within a $128 \times 128$ canvas.

We train Label2obj (Sec.~\ref{sec:label2obj}) with $7$ SPADE blocks, a mini-batch size of $16$, using Adam optimizer and $\lambda_D = \lambda_G = 10$. The initial learning rate is $2\times{10}^{-4}$ for $8$ epochs. The rate is then set to decay linearly  for next $4$ epochs. The output RGB image has spatial dimensions of $128 \times 128$.

\begin{figure}[!t]
  \centering
  \includegraphics[width=\linewidth]{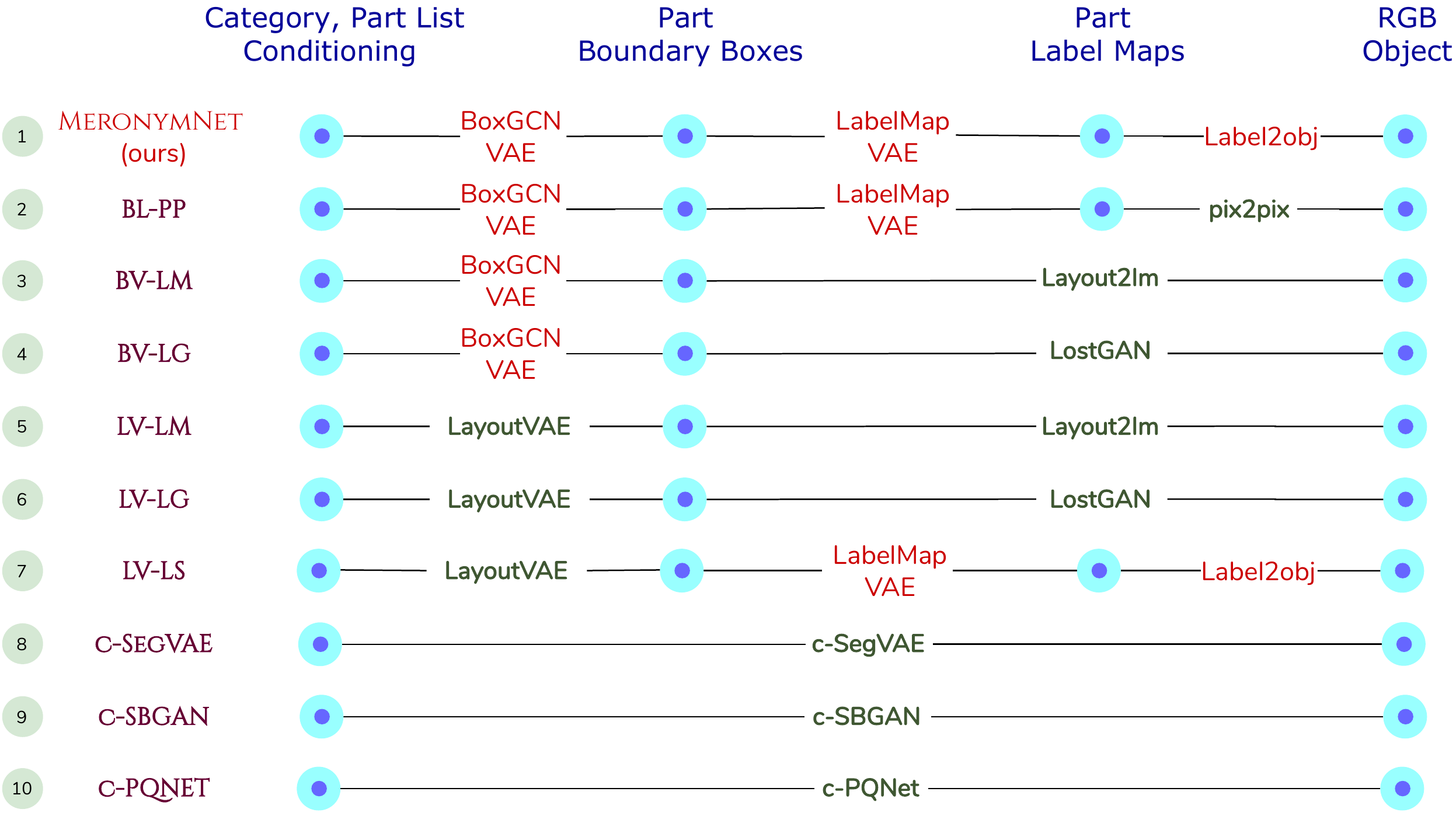}
  \caption{An illustration of baselines (rows) used for comparative evaluation against MeronymNet (top row). The blue concentric circles in columns denote inputs and outputs of components. Components reused from MeronymNet are shown in orange text and those based on our modifications to existing scene-based models are in green text.}
  \label{fig:baselines-list}
\end{figure}

\begin{figure*}[!ht]
  \centering
  \includegraphics[width=0.75\textwidth]{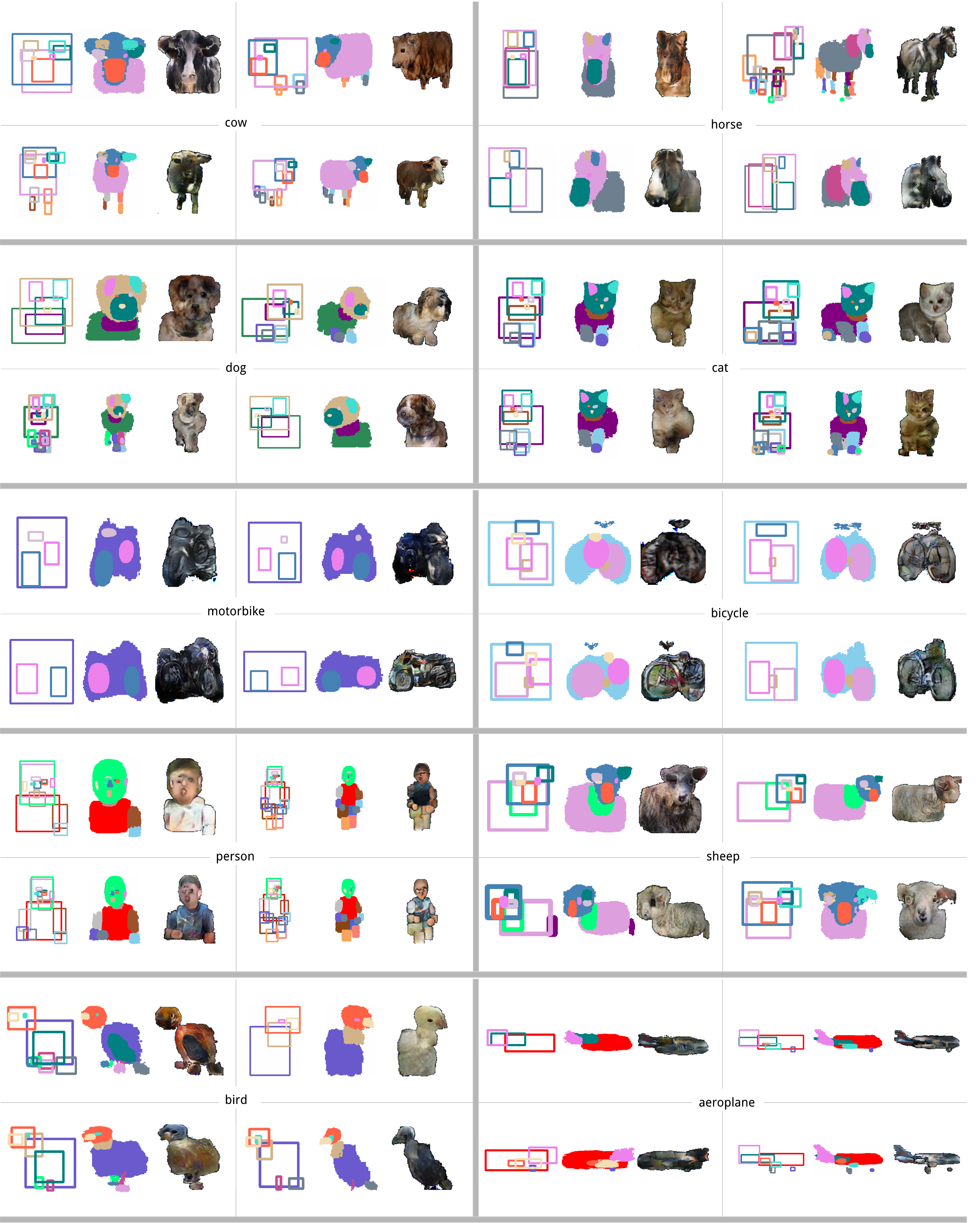}
  \caption{Each panel contains MeronymNet generations (part bounding box, label map, object) for each object category. Note that the generations are conditioned by object category and an associated list of parts (not shown to reduce clutter).}
  \label{fig:merogen}
\end{figure*}

\section{Experimental Setup}
\label{sec:experiments}

\noindent \textbf{Dataset:} For our experiments, we use the PASCAL-Part dataset~\cite{chen2014detect}, containing $10{,}103$ images across $20$ object categories annotated with part labels at pixel level. We select the following $10$ object categories: cow, bird, person, horse, sheep, cat, dog, airplane, bicycle, motorbike. The collection is characterized by a large diversity in appearance, viewpoints and associated part counts (see Fig.~\ref{fig:violin}). The individual objects are cropped from dataset images and centered. To augment images and associated part masks, we apply translation, anisotropic scaling for each part independently and also collectively at object level. We also augment via horizontal mirroring. The objects are then normalized with respect to the minimum and maximum width across all images such that all objects are centered in a $[0,1] \times [0,1]$ canvas. We use $75\%$ of the images for training, $15\%$ for validation and the remaining $10\%$ for evaluation.

\noindent \textbf{Baseline Generative Models:} Since no baseline models exist currently for direct comparison, we designed and implemented the baselines ourselves -- see Fig.~\ref{fig:baselines-list} for a visual illustration of baselines and component configurations. We modified existing scene layout generation approach~\cite{Jyothi_2019_ICCV} to generate part layouts. In some cases, we modified existing scene generation approaches having layouts as the starting point~\cite{Azadi2019SemanticBS,Sun_2019_ICCV,li2019layoutgan} to generate objects. We also included modified variants of two existing part-based object generative models (3-D objects~\cite{Wu_2020_CVPR}, faces~\cite{cheng2020segvae}). To evaluate individual components from MeronymNet, we also designed hybrid baselines with MeronymNet components (BoxGCN--VAE, LabelMapVAE, Label2obj) included. We incorporated object category, part-list based conditioning in each baseline to ensure fair and consistent comparison. Additional details regarding the baselines can be found in project page.

\noindent \textbf{Evaluation Protocol:} For each model (including MeronymNet), we generate $100$ objects for each category using the per-category part lists in the test set. We use Frechet Inception Distance~\cite{heusel2017gans} (FID) as a quantitative measure of generation quality. The smaller the FID, better the quality. For each model, we report FID for each category's generations separately and overall average FID.  

\begin{figure*}[!t]
  \centering
  \includegraphics[width=0.9\textwidth]{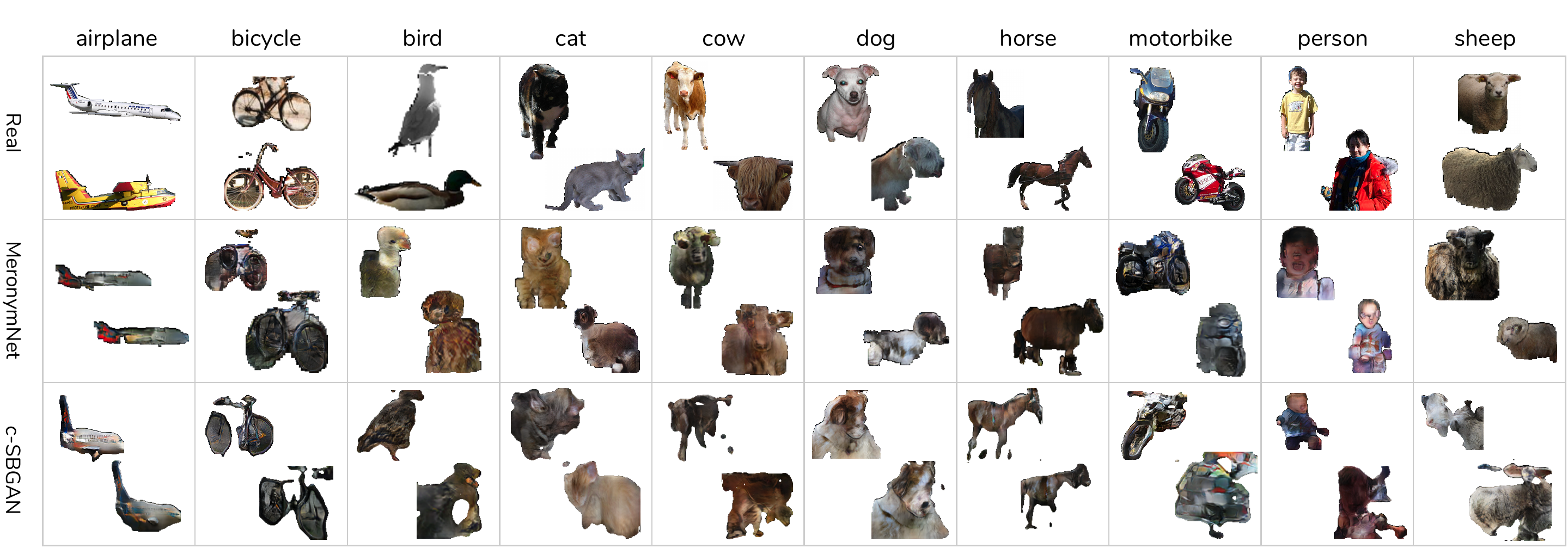}
  \caption{Sample object generations from MeronymNet (second row), c-SBGAN (third row) and objects from test set (top row).}
  \label{fig:qualitative-comparison}
\end{figure*}

\section{Results}
\label{sec:results}

\noindent \textbf{Qualitative results:} MeronymNet's generations obtained from object category and associated part lists from test set can be viewed in Fig.~\ref{fig:merogen}. The results demonstrate the MeronymNet's ability in generating diverse object maps for multiple object categories. The results also convey our model's ability to accommodate a large range of parts within and across object categories.  

\begin{table}[!t]
\resizebox{0.48\textwidth}{!}
{
\centering
\begin{tabular}{clc|cccccccccc}
\toprule
 \multicolumn{13}{c}{                  Frechet Inception Distance (FID) $\downarrow$} \\ 
\toprule
 \textbf{ID} & \textbf{Model} & \textbf{Overall} &  plane & bicycle & bird & cat & cow & dog & horse & m.bike & person & sheep \\
 \midrule
  \textbf{1} & \textsc{\textbf{MeronymNet}}  &  $\mathbf{331.6}$ & $341.9$   &  $384.1$   &  $340.3$  & $247.3$  & $334.1$  & $319.8$  & $364.5$  & $326.0$  & $368.5$ & $289.6$    \\
  {2} & \textsc{BL-PP}~\cite{isola2017image}  & $387.2$ &  $408.1$  & $413.0$ & $363.5$  & $315.9$ & $390.7$ & $385.7$  & $390.0$ & $454.5$ & $408.8$ & $341.6$ \\
  {3} & \textsc{BV-LM}~\cite{zhaobo2019layout2im}     & $397.3$  &  $415.6$ & $437.8$ & $346.4$ & $406.8$ & $371.5$ & $390.4$ & $370.4$ & $421.9$ & $420.6$ & $391.8$ \\
 {4} & \textsc{BV-LG}~\cite{Sun_2019_ICCV}     &    $368.9$  &  $421.7$  & $406.2$  & $335.3$  & $346.0$ & $338.9$  & $336.0$ & $337.5$ & $440.6$ & $393.1$ & $334.0$ \\
 {5} & \textsc{LV-LM}~\cite{Jyothi_2019_ICCV,zhaobo2019layout2im}     &    $405.6$  &  $419.2$ & $394.5$  & $374.5$  & $424.0$ & $386.4$  & $411.8$ & $387.6$  & $415.0$ & $456.0$ & $387.3$  \\
 {6} & \textsc{LV-LG}~\cite{Jyothi_2019_ICCV,Sun_2019_ICCV}     &    $420.8$  &  $426.2$ & $438.2$ & $389.2$ & $428.6$  & $390.2$ & $418.3$  & $402.6$ & $468.8$  & $438.3$ & $406.9$ \\
 {7} & \textsc{LV-LS}~\cite{Jyothi_2019_ICCV}     &    $373.8$  &  $361.3$ & $374.2$  & $351.2$  & $376.5$ & $383.6 $  & $366.7$ & $375.4$  & $385.0$ & $392.1$ & $372.3$ \\
 {8} & \textsc{c-SEGVAE}~\cite{cheng2020segvae}     &    $378.2$  &  $355.7$ & $367.1$  & $345.6$  & $406.9$ & $376.4$  & $402.7$ & $372.9$  & $379.5$ & $408.3$ & $367.4$ \\
 {9} & \textsc{c-SBGAN}~\cite{Azadi2019SemanticBS}     &    $342.4$   &  $287.7$  & $326.3$ & $321.0$ & $346.5$ & $350.3$ & $363.9$ & $356.8$  & $357.6$  & $379.4$ & $334.8$ \\
 {10} & \textsc{c-PQNET}~\cite{Wu_2020_CVPR}   &    $380.6$  &  $353.3$ & $335.1$  & $331.6$  & $410.0$ & $392.7$  & $385.7$ & $400.2$  & $368.7$ & $449.3$ & $379.4$ \\
\end{tabular}
 }
 \captionof{table}{Category-wise and overall FID for different baselines and MeronymNet. References to scene-generation components are provided alongside each baseline's name (column 2). Refer to Fig~\ref{fig:baselines-list} for a visual representation of baseline components.}
\label{tab:metrics}
\end{table}
\noindent \textbf{Quantitative results:} As Table~\ref{tab:metrics} shows, Meronymnet outperforms all other baselines. The quality of part box layouts from BoxGCN--VAE is better than those produced using LSTM-based LayoutVAE~\cite{Jyothi_2019_ICCV} (compare rows $1,7$, rows $3,5$ and $4,6$, also see Fig.~\ref{fig:baselines-list}). Note that the modified version of PQNet~\cite{Wu_2020_CVPR} (row $10$) also relies on a sequential (GRU) model. The benefit of modelling object layout distributions via the more natural graph-based representations (Sec.~\ref{sec:BoxGCN-VAE}) is evident.

\begin{table}[!t]
 
\centering
\resizebox{\linewidth}{!}
{
  \begin{tabular}{ c|c|l|c}
 \toprule
             Ablation Type & Pipeline Component & Ablation Details  & FID $\downarrow$ \\
 \midrule
\multirow[c]{5}{*}[-4.5em]{\rule{0pt}{2ex}Architectural} & \multirow[c]{4}{*}[-1.5em]{\rule{0pt}{2ex} BoxGCN--VAE (Sec.~\ref{sec:BoxGCN-VAE})} &  No skip connection & $348.7$  \\
& & No category conditioning in encoder & $ 366.7 $ \\ 
& & 1 GCN layer & $343.7$  \\
& & 3 GCN layers & $344.8$  \\
& & Reduce VAE latent dimensions (0.5x) & $343.1$  \\
& & Increase VAE latent dimensions (2x) & $334.5$  \\
\cline{2-4} 
& \multirow[c]{3}{*}[-1.5em]{\rule{0pt}{2ex} LabelMap--VAE (Sec.~\ref{sec:LabelMap-VAE}) } & No b.box conditioning in encoder & $391.7$ \\
& & No latent vector conditoning (${z}_{m}$) & $378.1$ \\
& & $128 \times 128$ part masks & $342.5$ \\
& & Reduce VAE latent dimensions (0.5x) & $336.4$  \\
& & Increase VAE latent dimensions (2x) & $340.2$  \\
\cline{2-4}
& \multirow[c]{2}{*}[-0.5em]{\rule{0pt}{2ex} Label2Obj (Sec.~\ref{sec:label2obj})} & No class conditioning  & $332.9$ \\
& &  $256 \times 256$ output resolution & $338.8$ \\
& &  With VAE & $335.4  $ \\
\cline{2-4}
& \multirow[c]{3}{*}[0.5em]{\rule{0pt}{2ex} Single level generation } & Generate b.boxes, masks and part RGB  (ref. project page) & $380.3$ \\
& & Generate b.boxes, combined masks and part RGB (ref. project page) & $395.0$ \\
\hline
\multirow{2}{*}[-1em]{\rule{0pt}{2ex} Optimization } & \multirow[c]{3}{*}{\rule{0pt}{2ex} BoxGCN--VAE (Eq.~\ref{eqn:BoxGCN-VAE-recons-loss})} & No IoU Loss & $348.6$  \\
& & No ${L}_{mn}^{mse-c}$ Loss & $348.4$  \\
& & No Adjacency Matrix Loss & $362.9$  \\
\cline{2-4}
& Overall & No constraint between train-val losses (Sec.~\ref{sec:impldetails}) & $380.4$  \\
\hline
\multicolumn{3}{c}{\textsc{MeronymNet}} & $\mathbf{331.6}$  \\
\bottomrule
\end{tabular}
}
\captionof{table}{Performance scores for ablative variants of MeronymNet.}
\label{tab:ablations}
\end{table}

The results also highlight the importance of our three-stage generative pipeline. In particular, MeronymNet distinctly outperforms approaches which generate objects directly from bounding box layouts~\cite{zhaobo2019layout2im,Sun_2019_ICCV} (compare rows $1,3,4$). Also, the relatively higher quality of MeronymNet's part label maps ensures better performance compared to other label map translation-based approaches~\cite{Azadi2019SemanticBS,cheng2020segvae,isola2017image} (compare rows $1,2,8,9$). In particular, note that our modified variants of some models~\cite{Azadi2019SemanticBS,cheng2020segvae} (rows $8,9$) employ a SPADE-based approach~\cite{park2019SPADE} for the final label-to-image stage. Model scores for another generative model quality measure - Activation Maximization Score~\cite{borji2019pros} - can be viewed in project page. The baseline rankings using AM Score also show trends similar to FID.

The quality of sample generations from MeronymNet (second row in Fig.~\ref{fig:qualitative-comparison}) is comparable to unseen PASCAL-Parts test images with same part list (top row). The same figure also shows sample generations from c-SBGAN, the next best performing baseline (last row). The visual quality of MeronymNet's generations is comparatively better, especially for categories containing many parts (e.g. person, cow, horse). Across the dataset, objects tend to have a large range in their part-level morphology (area, aspect ratio, count - Fig.~\ref{fig:violin}) which poses a challenge to image-level label map generation approaches, including c-SBGAN. In contrast, our design choices - generating all part label masks at same resolution (Fig.~\ref{fig:LabelMap-VAE}), decoupling bounding box and label map geometry -- all help address the challenges better~\cite{dornadula2019visual}. 

It is somewhat tempting to assume that parts are to objects what objects are to scenes, compositionally speaking. However, the analogy does not hold well in the generative setting. This is evident from our results with various scene generation models as baseline components (Table~\ref{tab:metrics}). The structural and photometric constraints for objects are stricter compared to scenes. In MeronymNet, these constraints are addressed by incorporating compositional structure and semantic guidance in a considered, coarse-to-fine manner which enables better generations and performance relative to baselines.

\noindent \textbf{Ablations:} We also compute the FID score for ablative variants of MeronymNet. The results in Table~\ref{tab:ablations} illustrate the importance of key design and optimization choices within our generative approach. In particular, note that the VAE version of Label2obj performs worse than our default (translation) model (Sec.~\ref{sec:label2obj}). Unlike scenes, object appearances do not depict as much diversity for a specified label map. This likely makes the VAE a more complex alternative during optimization.

\section{Interactive Modification}
\label{sec:interactive}

We now describe experiments used to obtain the results from the \textsc{MeroBot} app dialog mockup (Fig.~\ref{fig:merobot}). For the first scenario (panel $\mathbb{A}$), we generate an object based on the user-specified category (`cow'). If parts are not mentioned, a part list sampled from the category's training set distribution is chosen. If the user chooses to add new parts, the GUI enables canvas drawing editor controls. The user drawn part masks and corresponding labels are merged into the part label map generated earlier. When user clicks `Render' button, the updated label map is processed by Label2obj (Sec.~\ref{sec:label2obj}) to obtain the updated object generation. A similar process flow, but involving specification of bounding box based edits and BoxGCN--VAE (Sec.~\ref{sec:BoxGCN-VAE}) is used to generate a caricature-like exaggeration of the object depiction (panel $\mathbb{B}$). The third panel $\mathbb{C}$ depicts a process flow wherein the desired edits are specified at a high-level, i.e. in terms of part presence or absence. The updated part list is subsequently used for generating the corresponding object representations (last frame). Note that the viewpoint for rendering the object has changed from the initial generation to accommodate the updated part list. This scenario demonstrates MeronymNet's holistic, part-based awareness of rendering viewpoints best suited for various part sets.

\section{Conclusion}

In this paper, we have presented MeronymNet, a hierarchical and controllable generative framework for general collections of 2-D objects. Our model generates diverse looking RGB object sprites in a part-aware manner across multiple categories using a single unified architecture. The strict and implicit constraints between object parts, variety in layouts and extreme part articulations, generally make multi-category object generation a very challenging problem. Through our design choices involving GCNs, CVAEs, RNNs and guidance using object attribute conditioning, we have shown that these issues can be successfully tackled using a single unified model. Our evaluation establishes the quantitative and qualitative superiority of MeronymNet, overall and at individual component level. We have also shown MeronymNet's suitability for controllable object generation and interactive editing at various levels of structural and semantic granularity. The advantages of our hierarchical setup include efficient processing and scaling with inclusion of additional object categories in future. 

More broadly, our model paves the way for enabling compact, hierarchically configurable and truly end-to-end scene generation approaches. In this paradigm, the generation process can be controlled at class and part level to generate objects as we do and subsequently condition on these objects at the overall scene level. Going forward, we intend to explore modifications towards improved quality, diversity and degree of controllability. We also intend to explore the feasibility of our unified model for multi-category 3-D object generation. 

\bibliographystyle{ACM-Reference-Format}
\balance
\bibliography{main}

\end{document}